\documentclass[letterpaper]{article} 
\usepackage{aaai25}
\usepackage{times}  
\usepackage{helvet}  
\usepackage{courier}  
\usepackage[hyphens]{url}  
\usepackage{graphicx} 
\usepackage{natbib}  
\usepackage{caption} 
\usepackage{algorithm}
\usepackage{algorithmic}

\usepackage{multirow}
\usepackage{booktabs}
\usepackage{xcolor}
\usepackage{amssymb}
\usepackage{makecell}
\usepackage{amsmath}
\usepackage{subfig}
\usepackage{colortbl}
\usepackage{xspace}

\makeatletter
\DeclareRobustCommand\onedot{\futurelet\@let@token\@onedot}
\def\@onedot{\ifx\@let@token.\else.\null\fi\xspace}

\def\eg{\textit{e.g}\onedot} 
\def\ie{\textit{i.e}\onedot} 
\def\cf{\textit{c.f}\onedot}

\def\etal{\textit{et al}\onedot}
\makeatother

\usepackage{newfloat}
\usepackage{listings}
\DeclareCaptionStyle{ruled}{labelfont=normalfont,labelsep=colon,strut=off} 
\floatstyle{ruled}
\newfloat{listing}{tb}{lst}{}
\floatname{listing}{Listing}
\title{VidCompress: Memory-Enhanced Temporal Compression for \\ Video Understanding in Large Language Models}
\author{
    Xiaohan Lan,
    Yitian Yuan,
    Zequn Jie,
    Lin Ma
}
\affiliations{
    Meituan Inc.
}

\begin{document}

\maketitle

\begin{abstract}
Video-based multimodal large language models (Video-LLMs) possess significant potential for video understanding tasks. However, most Video-LLMs treat videos as a sequential set of individual frames, which results in insufficient temporal-spatial interaction that hinders fine-grained comprehension and difficulty in processing longer videos due to limited visual token capacity. To address these challenges, we propose VidCompress, a novel Video-LLM featuring memory-enhanced temporal compression. VidCompress employs a dual-compressor approach: a memory-enhanced compressor captures both short-term and long-term temporal relationships in videos and compresses the visual tokens using a multiscale transformer with a memory-cache mechanism, while a text-perceived compressor generates condensed visual tokens by utilizing Q-Former and integrating temporal contexts into query embeddings with cross attention. Experiments on several VideoQA datasets and comprehensive benchmarks demonstrate that VidCompress efficiently models complex temporal-spatial relations and significantly outperforms existing Video-LLMs.

\end{abstract}

%

\section{Introduction}

Large Language Models (LLMs) have gained significant attention in the field of artificial intelligence (AI) due to their remarkable capabilities in understanding and generating human language. Models like GPT-3.5~\cite{openai2023chatgpt}, GPT-4~\cite{achiam2023gpt} and LLaMA-3~\cite{dubey2024llama} demonstrate impressive performance across various natural language tasks like text generation, sentiment analysis, and machine translation~\cite{devlin-etal-2019-bert}. Based on the powerful language and knowledge capabilities of LLMs, some studies~\cite{li2023blip, zhu2023minigpt} resort to extend text-only understanding by converting the visual input signals, such as images and videos, into tokens that LLMs can understand. Such multimodal LLMs can take more modalities as inputs, significantly broadening the applications of LLMs in AI communities to comprehend the diverse aspects of the physical world.

\begin{figure}[!t]
	\centering
	\includegraphics[width=\columnwidth]{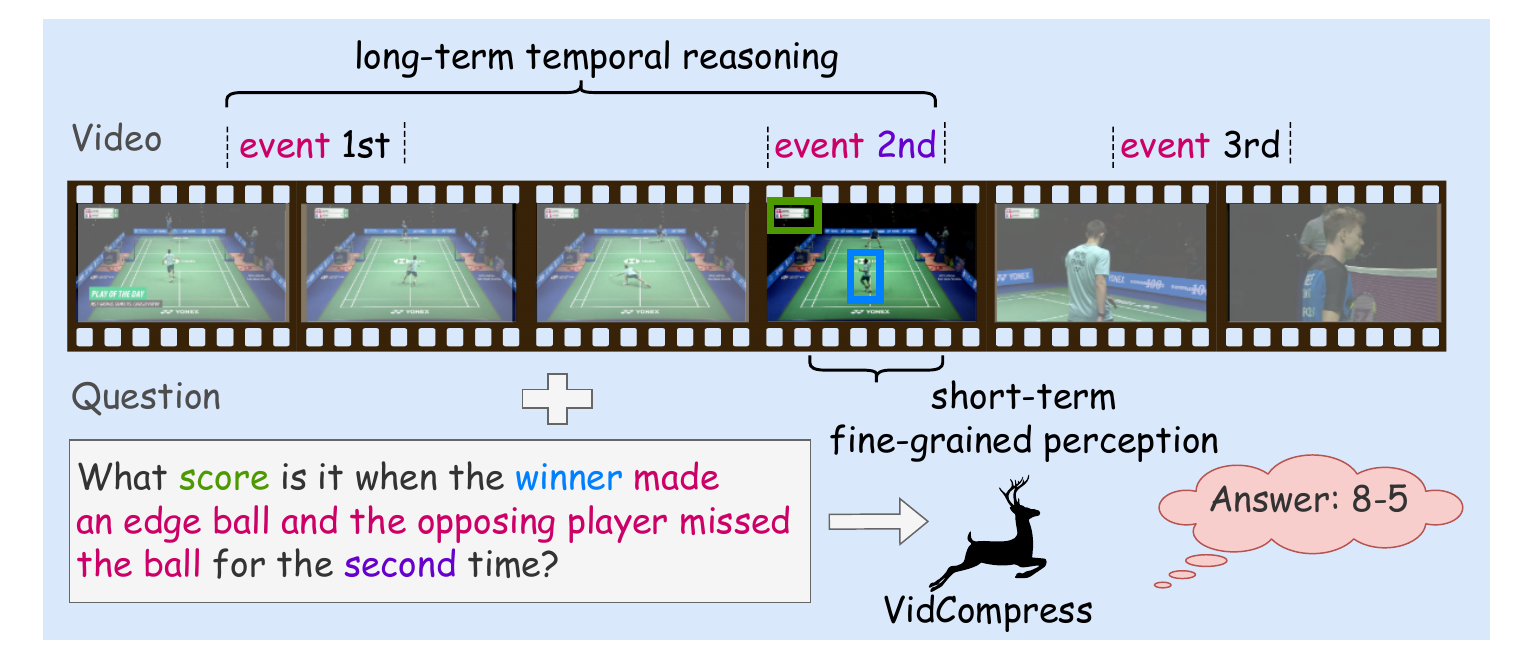}
	\caption{An example of a badminton match video, where temporal reasoning is required to detect the \textcolor[HTML]{CC0066}{event} of the \textcolor[HTML]{007FFF}{winner} scoring a point with mading a second edge ball, and single-frame fine-grained recognition is also needed to identify the specific \textcolor[HTML]{4D9900}{score}. Our proposed VidCompress, with a memory-aware dual-compressor architecture, is capable of performing both long-term and short-term temporal modeling to correctly answer the question.
	}
	\label{fig:example}
\end{figure}

For multimodal large language models, to comprehend videos takes more challenges than images. The majority of recent Video-LLMs~\cite{li2023llama, lin2023video} do not take the video as a whole but view it as a set of images. In other words, the video is transformed into sequential visual tokens using an image-level encoder followed by an adapter/projector. Temporal reasoning in Large Language Models (LLMs) is exclusively facilitated by the attention mechanisms within the transformer blocks. This way of indirect and late temporal modeling has two inevitable drawbacks: Firstly, changes between consecutive frames in videos convey motion or specific actions, representing short-term correlations. Meanwhile, frames farther apart depict logical relationships within events, indicating long-term associations. Relying solely on the LLM to manage the relationships between visual tokens in the video falls short of achieving such precise modeling, resulting in deficiencies in understanding fine-grained objects or actions, as well as in capturing enduring event connections. Second, the number of visual tokens input into the LLM increases with the length of the video, while the feasible visual token amount fed into the LLM is limited. Therefore, how to use limited visual token capacity to represent the informative video with complex temporal-spatial object relations for Video-LLMs becomes an essential issue to be addressed.

To better model temporal short-term correlations and long-term associations across frames while keeping efficiency, as illustrated by the example in Figure~\ref{fig:example}, we propose a novel Video-LLM named VidCompress, which enables memory-enhanced temporal compression in videos. VidCompress employs a dual-compressor architecture, consisting of a memory-enhanced compressor and a text-perceived compressor, to transform input videos into two types of visual tokens. 

Specifically, we segment the entire video into fixed-size clips and sequentially feed them into the memory-enhanced compressor to generate memory visual tokens. The memory-enhanced compressor is composed of a multiscale transformer with memory-cache mechanism~\cite{wu2022memvit}. Within a clip, the transformer performs inter-frame interactions by its spatial-temporal attention, aggregating temporal-adjacent information to build the short-term correlations. Also, benefited from the devised memory mechanism, the contextual information from previous clips can be preserved to model the long-term associations. 
In addition to long-/short-term temporal modeling, fine-grained perception within static frames is crucial for comprehensive video understanding as well. To this end, we introduce a text-perceived compressor to produce perceived visual tokens. First, we leverage Q-Former to compress the frame-wise visual feature, maintaining instruction-relevant visual contents. A cross-attention module is then employed to further integrate temporal contexts, using the memory-enhanced visual tokens from the other compressor as queries to create more condensed text-perceived visual tokens.

Afterwards, both the memory-enhanced tokens and text-perceived tokens of videos are adapted and input into the LLM alongside textual instructions to yield predicted textual tokens. To fully exploit the potential temporal reasoning power of VidCompress, we design a progressive training paradigm, encompassing both modality alignment and instruction tuning stages. We conduct experiments on multiple video question answering (VideoQA) datasets and benchmarks. Both experimental results and qualitative cases demonstrate that VidCompress excels in capturing temporal relationships as the video length increases. To conclude, our contributions can be summarized as follows:

\begin{itemize}
    \item We introduce VidCompress, a novel Video-LLM that employs a dual-compressor architecture. This architecture integrates a memory-enhanced compressor and a text-perceived compressor to effectively transform input videos into two distinct types of visual tokens, thereby enhancing temporal understanding.
    \item Our memory-enhanced compressor employs a multiscale transformer with a memory-cache mechanism to capture both short-term and long-term temporal relationships, enhancing video comprehension.
    \item Experiments show VidCompress has achieved promising results on VideoQA tasks and multiple Video-LLM benchmarks, highlighting the potential of introducing early temporal modeling for Video-LLMs.
\end{itemize}

\section{Related Work}
In this section, we review recent research on LLMs and multimodal LLMs, as well as advancements in long-form video understanding.

\paragraph{Large Language Models}
By extending the scale of both data and model parameters, we ushered in a brand-new era of large language models. Based on the transformer architecture, a series of language foundation models with billionaire-level parameters such as LLaMA~\cite{touvron2023llama}, GPT~\cite{achiam2023gpt} and Claude~\cite{anthropic2024claude} have emerged with powerful language reasoning and conversational ability after training on large-scale data. More open-source models such as Alpaca~\cite{taori2023stanford} and Vicuna~\cite{chiang2023vicuna} leverage the strategy of instruction tuning to further improve the foundation model. Given the powerful tokenized textual understanding and generalization ability, we can build a multimodal large language model via tokenizing the visual signals.

\paragraph{Multimodal Large Language Models}

Adapting LLMs to interpret visual tokens, multimodal large language models (MLLMs) can process both textual and visual inputs and produce coherent responses. Typically, MLLMs bridge the vision-language gap with lightweight adapters. For example, LLaVA~\cite{liu2023llava,liu2023improvedllava} and MiniGPT-4~\cite{zhu2023minigpt} use a linear layer to project visual features into the language hidden space, while BLIP/BLIP-2~\cite{li2022blip, li2023blip} introduces a query transformer (Q-Former) to efficiently extract visual features using learnable query embeddings. Additionally, high-quality image-text instruction pairs~\cite{chen2023sharegpt4v} are created for multimodal pretraining and fine-tuning to align the textual-visual space.

To support video understanding, researchers extend image-based MLLMs for video inputs. For instance, mplug-owl~\cite{ye2023mplug} processes video inputs similarly to images, which might overlook inter-frame dependencies. To address this, Video-ChatGPT~\cite{maaz2023video} incorporates pooling to extract spatial and temporal features, while VideoChat~\cite{li2023videochat} adds a temporal modeling module between the visual encoder and adapter. Other methods use video-based visual encoders, such as VideoLLaVA~\cite{lin2023video} which initializes the multimodal encoder from LanguageBind~\cite{zhu2023languagebind}, and VideoChat2, which employs UMT~\cite{li2023unmasked} for feature extraction. However, as the video length increases, the number of visual tokens also grows, leading to a longer context for LLMs to process with reduced efficiency. Some approaches mitigate this by downsampling frames, fixing video token lengths, or reducing tokens per frame. For instance, MovieChat~\cite{song2023moviechat} employs a long-short memory mechanism for global or breakpoint understanding, while LLaMA-VID~\cite{li2023llama} reduces visual tokens by representing each frame with two tokens.

\begin{figure*}[!thb]
	\centering
	\includegraphics[width=0.9\textwidth]{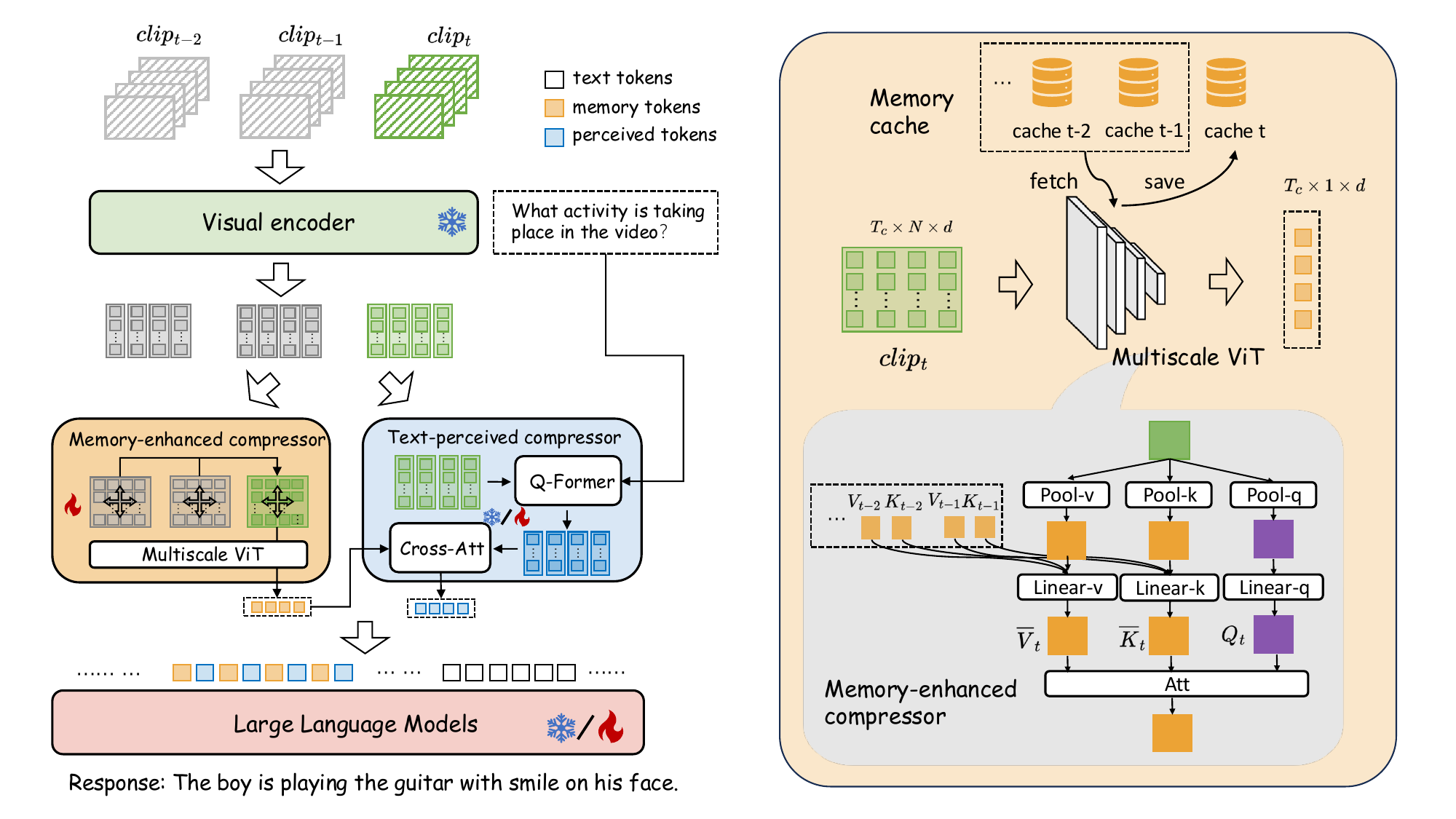}
	\caption{The overall framework of our proposed VidCompress, following a dual-compressor architecture. The visual encoder extracts frame-level features that are fed into the memory-enhanced compressor and text-perceived compressor to generate two types of visual tokens. The right part details the memory-enhanced compressor with devised memory-cache strategy.
	}
	\label{fig:intro}
\end{figure*}

\paragraph{Long-form Video Understanding}
Long-form video understanding is a traditional yet challenging problem in visual perception due to the high computational cost and complexity of modeling temporal relationships among lengthy video frames. Some approaches~\cite{li2017temporal, miech2017learnable} pre-compute visual features with a frozen backbone to reduce training overhead, while others use sparse sampling, which leads to information loss. Alternative methods~\cite{wang2016temporal, christoph2016spatiotemporal} use cache/memory mechanisms to handle long video sequences. Wu~\etal~\citeyear{wu2019long} introduce a long-term feature bank for detailed video understanding, and MeMViT~\cite{wu2022memvit} adds an augmented memory module to the transformer-based MViT~\cite{fan2021multiscale} for more efficient video length scaling. Inspired by MeMViT, our approach also employs a memory-cached strategy within transformer blocks to integrate informative semantics from previous frames to current ones.


\section{Method}
This section offers a comprehensive description of our proposed VidCompress, detailing each module and the associated training strategy.
\subsection{Overview}
As illustrated in Figure~\ref{fig:intro}, the framework consists of serveral key components, \ie, visual encoder, memory-enhanced compressor, text-perceived compressor and token adaption process for the LLM.

\subsection{Visual Encoder}
The visual encoder is responsible for extracting visual features from the input video frames. It processes a sequence of $T$ frames, utilizing a vision transformer (ViT) to transform each frame into several visual tokens/patches. Therefore, the encoded video feature is formulated as $\mathbf{F}_v = \{\mathbf{v}_1, \mathbf{v}_2, ..., \mathbf{v}_T\}, \mathbf{v}_i \in \mathbb{R}^{N \times d}$, where $N$ is the number of tokens/patches within each frame.

\subsection{Memory-enhanced Compressor}
Inspired by recent advances of video foundation models, we design the memory-enhanced compressor following the MeMViT~\cite{wu2022memvit} architecture. MeMViT, or Memory-Augmented Multiscale Vision Transformer, is a state-of-the-art model designed for efficient long-term video recognition. Building upon the Multiscale Vision Transformers (MViT) framework, MeMViT introduces memory augmentation to enhance its ability to capture and retain long-term dependencies in video sequences, which is essential for understanding activities that unfold over extended periods. 

After obtaining $\mathbf{F}_v \in \mathbb{R}^{T \times N \times d}$ from the visual encoder, we further split it by clips with a fixed size. For the $t$-th video clip,  its clip feature can be represented as $\mathbf{C}_t \in \mathbb{R}^{T_c \times N \times d}$, where $T_c$ denotes the clip size. Then, we sequentially feed these clip features into the memory-enhanced compressor to generate memory-enhanced visual tokens.
As shown in the right part of Figure~\ref{fig:intro},  the $t$-th video clip feature $\mathbf{C}_t$ is fed into a four-layer multiscale transformer blocks, and each block performs spatial downsampling through a pooling attention operation with memory cache interaction. For one transformer block, we first take its input $\mathbf{X}_t$~(the first block's input is $\mathbf{C}_t$) to a pooling layer:
\begin{align}
    \mathbf{Q}_t &= \mathcal{P}(\mathbf{W}_{Q}\mathbf{X}_t)\,,\\
    \mathbf{K}_t &= \mathcal{P}(\mathbf{W}_{K}\mathbf{X}_t)\,, \\
    \mathbf{V}_t &= \mathcal{P}(\mathbf{W}_{V}\mathbf{X}_t)\,. 
\end{align}
Here, $\mathcal{P}$ performs a 3D convolution operation with 3D stride $(s_T,s_H,s_W)$, kernel $(k_T,k_H,k_W)$, and padding $(p_T,p_H,p_W)$. By setting appropriate stride, kernel and padding numbers, we could downsample the input's spatial dimension while keeping the temporal dimension. In this paper, each transformer block will downsample the input's spatial dimension by 4, and therefore after four blocks' computation, the $N=256$ patch-level visual tokens of a single frame will be compressed into one unified token.

After the pooling layer, we further extend $\mathbf{K}_t$ and $\mathbf{V}_t$ with cached memory of the K and V values from previous clips and perform attention with these augmented keys and values:
\begin{align}
    \overline{\mathbf{K}_t} &= \text{Concat}\left[\mathbf{K}^{\text{cache}}, \mathbf{K}_t\right]\,, \\
    \overline{\mathbf{V}_t} &= \text{Concat}\left[\mathbf{V}^{\text{cache}}, \mathbf{V}_t\right]\,, \\
    \mathbf{Z}_t &= \text{Attn}(f_{\text{linear}_Q}(\mathbf{Q}_t) \nonumber \\
    &f_{\text{linear}_K}(\overline{\mathbf{K}_t}), f_{\text{linear}_V}(\overline{\mathbf{V}_t}))\,.
\end{align}
Here $\mathbf{K}^{\text{cache}}$ and $\mathbf{V}^{\text{cache}}$ are the key and value vectors of previous $M$ video clips, which is pre-stored in the memory cache. Thus, when we conduct attention operation, each video clip could interact with previous clips, and thus the long-range temporal context can be retained in this procedure. Notably, the spatial and temporal dimensions of all the Q/K/V values are flattened, thus such spatial-temporal attention also facilitates inter-frame fine-grained interactions. Finally, the output $\mathbf{Z}_t$ will be taken as the input to the next transformer block, and get further token compression.

To this end, both the long-term and short-term correlation could be established in this memory-enhanced attention mechanism, which is crucial for understanding activities spread over long video sequences. Additionally, since our video clips can be processed sequentially, each video clip only needs to access the previous $M$ clips when computing attention. Therefore, the memory mechanism can be implemented in a first-in-first-out queue manner.

After the memory-enhanced compressor, the input video feature $\mathbf{F}_v \in \mathbb{R}^{T \times N \times d}$ are compressed in the spatial dimension, and thus yielding the memory-enhanced visual tokens $\mathbf{F}_m \in \mathbb{R}^{T \times 1 \times d}$.

\subsection{Text-perceived Compressor}

In addition to long-term and short-term temporal modeling, the text-perceived compressor concentrates more on intra-frame interactions under the text/question guidance. It generates text-perceived visual tokens by utilizing Q-Former and a devised cross-attention module in a two-step compression manner. 

At the first step, by employing the text-aware Q-Former similar to InstructBLIP~\cite{instructblip}, the visual tokens of each video frame (\eg, $\mathbf{v}_i$) are compressed into $N_q$ query tokens individually, thus maintaining text-relevant visual contents. For the $i$-th frame, we define the output of text-aware Q-Former as $\mathbf{q}_i \in \mathbb{R}^{N_q \times d}$. At the second step, to incorporate temporal context information into text-aware visual tokens $\mathbf{q}_i$, we employ a cross-attention module that aggregates both temporal and textual context and further compress $\mathbf{q}_i$ into one compact token. More specifically, for $i$-th frame, we take its corresponding memory-enhanced visual tokens $\mathbf{F}_{m}[i] \in \mathbb{R}^{1\times d}$ from the memory-enhanced compressor as the query to attend $\mathbf{q}_i$ and obtain its final compressed text-perceived visual tokens $\mathbf{F}_{p}[i]$:
\begin{align}
    \mathbf{F}_{p}[i] &= \text{CrossAttn}(\mathbf{F}_{m}[i], \mathbf{q}_i, \mathbf{q}_i) \nonumber \\
    &= \text{Softmax}(\mathbf{F}_{m}[i] \cdot \mathbf{q}_i^T / \sqrt{d}) \cdot \mathbf{q}_i
    \,.
\end{align}
By aggregating all frames' text-perceived visual tokens, we thus could get $\mathbf{F}_p \in \mathbb{R}^{T \times 1 \times d}$.


\begin{table}[!t]
  \centering
    \begin{tabular}{l|cc}
    \toprule
    \multirow{2}[2]{*}{Settings} & \multicolumn{2}{c}{Training Phase} \\
          & \multicolumn{1}{p{5.585em}}{Modality Alignment} & \multicolumn{1}{p{5.585em}}{Instruction Tuning} \\
    \midrule
    Batch Size & 256   & 128 \\
    Epoch & \multicolumn{2}{c}{1} \\
    Learning Rate & 1e-3  & 2e-5 \\
    Learning Schedule & \multicolumn{2}{c}{Cosine Decay} \\
    Warmup Ratio & \multicolumn{2}{c}{0.03} \\
    Weight Decay & \multicolumn{2}{c}{0} \\
    Optimizer & \multicolumn{2}{c}{AdamW} \\
    Max Token & \multicolumn{2}{c}{2048} \\
    Visual Encoder & \multicolumn{2}{c}{Freeze} \\
    Mem-enhanced Comp. & \multicolumn{2}{c}{Open} \\
    Q-Former & Freeze & Open \\
    Projectors & \multicolumn{2}{c}{Open} \\
    LLM   & Freeze & Open \\
    Video FPS & \multicolumn{2}{c}{1} \\
    \bottomrule
    \end{tabular}%
    \caption{Training parameter settings of VidCompress, \textit{Mem-enhanced Comp.} denotes the memory-enhanced compressor.}
  \label{tab:imple_details}%
\end{table}%

\subsection{Token Adaption for LLM}
As shown in Figure~\ref{fig:intro}, the outputs of both memory-enhanced compressor $\mathbf{F}_m$, and the outputs of the text-perceived compressor $\mathbf{F}_p$ are then adapted to the language semantic space with two linear projectors. Thus, we get the final adapted memory and perceived visual tokens to represent the input video. Then, all the visual tokens along with the language tokens from the input text are fed into the pretrained Language Foundation Model (LLM) to return a reasonable video-based response. 

In summary, our proposed VidCompress effectively incorporates multimodal context information. The use of dual-compressor architecture ensures that the model captures both temporal and language context to comprehend the video contents in an efficient way.

\begin{table*}[htbp]
  \centering
    \begin{tabular}{l|lccccc}
    \toprule
    Model Name & LLM   & \multicolumn{1}{l}{Res.} & \multicolumn{1}{l}{MVBench} & \multicolumn{1}{l}{Video-MME} & \multicolumn{1}{l}{LVBench} & \multicolumn{1}{l}{MMBench-Video} \\
    \midrule
    GPT-4V~\cite{openai2023gpt4v} & - & 224   & 43.70  & -     & -     & 1.53  \\
    GPT-4o~\cite{openai2024gpt4o} & - & 224   & -     & -     & 27.00  & 1.30  \\
    Gemini-1.5-Pro~\cite{reid2024gemini} & - & 224   & -     & -     & 33.10  & 1.44  \\
    \midrule
    VideoChat2~\cite{li2023mvbench} & Vicuna-7B & 224   & 51.10  & 39.50  & -     & 0.99  \\
    Video-LLaVA~\cite{lin2023video} & Vicuna-7B & 224   & -     & 39.90  & -     & - \\
    TimeChat~\cite{ren2023timechat} & LLaMA2-7B & 224   & -     & -     & 22.30  & - \\
    PLLaVA~\cite{xu2024pllava} & Vicuna-7B & 224   & 46.60  & -     & -     & 1.03  \\
    ShareGPT4Video~\cite{chen2024sharegpt4video} & LLaMA3-8B & 224   & \textbf{51.20}  & -     & -     & 1.05  \\
    \midrule
    \rowcolor{gray!15}
    VidCompress~(Ours) & Vicuna-7B & 224   & 46.85  & \textbf{43.00}  & \textbf{28.68}  & \textbf{1.14}  \\
    \bottomrule
    \end{tabular}%
    \caption{The comparison of Video-LLMs on different video benchmarks. The metric for Video-MME represents the overall score training without subtitles. ``-" denotes the value is not accessible. \textbf{Bold} indicates the best among open-source models.}
  \label{tab:videobenchmark}%
\end{table*}%

\subsection{Training Strategy}
As shown in Figure~\ref{fig:intro}, we freeze/unfreeze some modules for different training stages. Generally, we divide the whole training procedure into two stages, namely modality alignment and instruction tuning, respectively. 

For the modality alignment stage, we follow LLaMA-VID~\cite{li2023llama} and use 790K high-quality image-text and video-text pairs to pretrain our VidCompress model. In this stage, the model mainly focuses on the alignment of vision and language semantic space. Therefore, we only unfreeze the memory-augmented compressor, the cross-attention module, and the two projectors involved in the token adaption. Other modules like visual encoder and Q-Former are frozen during this stage.

For the instruction tuning stage, the model should be fully trained for comprehensive video understanding and instruction following. To this end, we further unfreeze the Q-Former and language foundation model besides the unfreezed modules in the previous stage. We build our instruction tuning dataset from two sources. One part includes 763K pure-text/image/video QA pairs collected by LLaMA-VID. To prove the temporal reasoning ability, we also include 230K video QA pairs sampled from VideoChat2~\cite{li2023mvbench}. 

\begin{table*}[htbp]
  \centering
    \begin{tabular}{l|lccccccc}
    \toprule
    \multirow{2}[2]{*}{Model Name} & \multirow{2}[2]{*}{LLM} & \multirow{2}[2]{*}{Res.} & \multicolumn{2}{c}{MSVD\mbox{-}QA} & \multicolumn{2}{c}{MSRVTT\mbox{-}QA} & \multicolumn{2}{c}{ActivityNet\mbox{-}QA} \\
          &       &       & Acc   & Score & Acc   & Score & Acc   & Score \\
    \midrule
    VideoLLaMA~\cite{zhang2023video} & DeBERTa\mbox{-}V2 & 224   & 51.6  & 2.5   & 29.6  & 1.8   & 12.4  & 1.1  \\
    LLaMA\mbox{-}Adapter~\cite{zhang2023llama} & Vicuna\mbox{-}7B & 224   & 54.9  & 3.1   & 43.8  & 2.7   & 34.2  & 2.7  \\
    VideoChat~\cite{li2023videochat} & LLaMA\mbox{-}7B & 224   & 56.3  & 2.8   & 45.0  & 2.5   & 26.5  & 2.2  \\
    VideoChat2~\cite{li2023mvbench} & Vicuna\mbox{-}7B & 224   & 70.0  & 3.9   & 54.1  & 3.3   & \textbf{49.1}  & 3.3  \\
    Video\mbox{-}ChatGPT~\cite{maaz2023video} & Vicuna\mbox{-}7B & 224   & 64.9  & 3.3   & 49.3  & 2.8   & 35.2  & 2.7  \\
    BT\mbox{-}Adapter~\cite{liu2023one} & Vicuna\mbox{-}7B & - & 67.5  & 3.7   & 57.0  & 3.2   & 45.7  & 3.2  \\
    Video\mbox{-}LLaVA~\cite{lin2023video} & Vicuna\mbox{-}7B & 224   & \textbf{70.7}  & 3.9   & \textbf{59.2}  & 3.5   & 45.3  & 3.3  \\
    LLaMA\mbox{-}VID~\cite{li2023llama} & Vicuna\mbox{-}7B & 224   & 69.7  & 3.7   & 57.7  & 3.2   & 47.4  & 3.3  \\
    \midrule
    \rowcolor{gray!15}
    VidCompress~(Ours) & Vicuna\mbox{-}7B & 224   & 68.9  & 3.7   & 57.7  & 3.2   & 48.3  & 3.3  \\
    \bottomrule
    \end{tabular}%
    \caption{The comparison of Video-LLMs on VideoQA datasets, with metrics of accuracy(\%) and average GPT-evaluated scores.}
  \label{tab:zs_videoqa}%
\end{table*}%

\section{Experiments}
In this section, we present the experimental setup and benchmark VidCompress against other leading Video-LLMs. Also, we analyze key components and provide qualitative results.

\subsection{Implementation Details}


We adopt ViT-G/14 from EVA-G~\cite{fang2023eva} as the visual encoder and Vicuna-7B~\cite{chiang2023vicuna} as the LLM. Q-Former for the text-perceived compressor are initialized by its pretrained weights~\cite{li2023blip}. Our memory-enhanced compressor refers to the structure of MeMViT but does not use the pretrained checkpoints. Instead, we design a customized, lightweight model with a 4-layer transformer trained from scratch. In our setup, the memory-enhanced compressor is connected to the back of the ViT to handle token compression. To keep the model lightweight and suitable for training, we set the video clip size to 8 and the cached memory size $M$ to 3, where the values to choose would be discussed in ablation studies. Furthermore, we configure the stride, kernel and padding of the 3D convolution to (1, 2, 2), (3, 3, 3) and (2, 2, 2), respectively, enabling each transformer layer to perform 4x spatial downsamping while preserving the temporal dimension. More training settings are depicted in Table~\ref{tab:imple_details}. The whole training procedure costs 72 hours on 8 A100 GPUs.

\subsection{Results on Video Benchmarks}


\begin{table*}[htbp]
\centering
    \begin{tabular}{l|cc|ccccc}
    \toprule
    \multirow{2}[2]{*}{Model} & \multirow{2}[2]{*}{\makecell[c]{memory \\ token}} & \multirow{2}[2]{*}{\makecell[c]{perceived \\ token}} & \multicolumn{4}{c}{Video-MME} & \multirow{2}[2]{*}{MVBench} \\
          &       &       & short & medium & long  & all   &  \\
    \midrule
    VidCompress$_{\rm{mem}}$  & \checkmark &       & 42.5  & 37.8  & 33.2  & 38.3  & 41.3  \\
    VidCompress$_{\rm{txt}}$ &       & \checkmark & \textbf{48.6}  & \textbf{41.9}  & 37.2  & \textbf{43.1}  & 45.6  \\
    \rowcolor{gray!15}
    VidCompress$_{\rm{full}}$ & \checkmark & \checkmark & 46.4  & \textbf{41.9}  & \textbf{37.8}  & 43.0  & \textbf{46.9}  \\
    \bottomrule
    \end{tabular}%
      \caption{Ablation studies on VidCompress branches. We show the results on Video-MME and MVBench.}
  \label{tab:ablation}%
\end{table*}%

We compare our VidCompress with other state-of-the-art models on several video benchmarks, including MVBench~\cite{li2023mvbench}, Video-MME~\cite{fu2024video}, LVBench~\cite{wang2024lvbench}, and MMBench-Video~\cite{fang2024mmbench} (\cf, Table~\ref{tab:videobenchmark}). Among these benchmarks, MVBench contains shorter videos, Video-MME includes short, medium, and long videos, while LVBench and MMBench-Video consist of longer videos. For relative fairness, we selected models that utilize 7B/8B LLMs for comparisons. Most of the open-source models also utilize Vicuna-7B as their LLMs and operate at an image resolution of 224, except for TimeChat and ShareGPT4Video, which employ different versions of the LLaMA model.

In summary, VidCompress demonstrates superior performance across multiple video benchmarks, outperforming comparable models such as VideoChat2, Video-LLaVA and PLLaVA. Meanwhile, our VidCompress shows particular strengths on the Video-MME, MMBench-Video and LVBench benchmarks that have longer videos, indicating its robust capability in comprehending complex video scenarios. This demonstrates the superiority of our video token compression mechanism and memory-aware temporal modeling design in analyzing and processing long videos. Specifically, on the Video-MME benchmark that includes minute-/hour-level testing videos, VidCompress achieves around 3\% performance higher than VideoChat2 and Video-LLaVA. For another long-video benchmark LVBench, VidCompress outperforms TimeChat with 6.38\% and the result is also competitive with closed-source methods like GPT-4o and Gemini-1.5-Pro. Besides, VidCompress achieves a score of 1.14 on MMBench-Video, surpassing VideoChat2, PLLaVA and ShareGPT4Video with significant gains. 

On the benchmark with relatively shorter videos, our VidCompress also achieves comparable results with other models. The accuracy of VidCompress on MVBench is 46.85, placing it close to PLLaVA of 46.6 and slightly lower than ShareGPT4Video. The reason for not performing so well is due to that the short video is not able to benefit from our token compression strategy with the devised memory mechanism, so that the temporal reasoning capability is withheld.


\subsection{Results on Video QA datasets}
We also compare VidCompress with other state-of-the-art models on several video question-answering (Video-QA) datasets, including MSVD-QA~\cite{chen2011collecting}, MSRVTT-QA~\cite{xu2016msr}, and ActivityNet-QA~\cite{caba2015activitynet}~(\cf, Table~\ref{tab:zs_videoqa}). VidCompress consistently performs at a high level across all three datasets, demonstrating its effectiveness in completing video-based QA tasks. More specifically, VidCompress shows strong performance in the MSVD-QA benchmark with an accuracy of 68.9\% and a score of 3.7, placing it among the top models, slightly behind Video-LLaVA and LLaMA-VID. In the MSRVTT-QA benchmark, VidCompress achieves an accuracy of 57.7\% and a score of 3.2, which is competitive with other leading models like LLaMA-VID and BT-Adapter. On the ActivityNet-QA benchmark, VidCompress achieves an accuracy of 48.3\% and a highest score of 3.3, which is comparable to LLaMA-VID and Video-LLaVA, both of which score 3.3 as well. Among these three datasets, ActivityNet-QA contains longest videos with diverse human activities. The superior results on ActivityNet-QA highlight VidCompress's capability to handle complex Video-QA tasks, making it a strong contender in the field.

\subsection{Ablation Studies on VidCompress Branches}


\begin{figure*}[tbhp]
	\centering
	\includegraphics[width=0.76\textwidth]{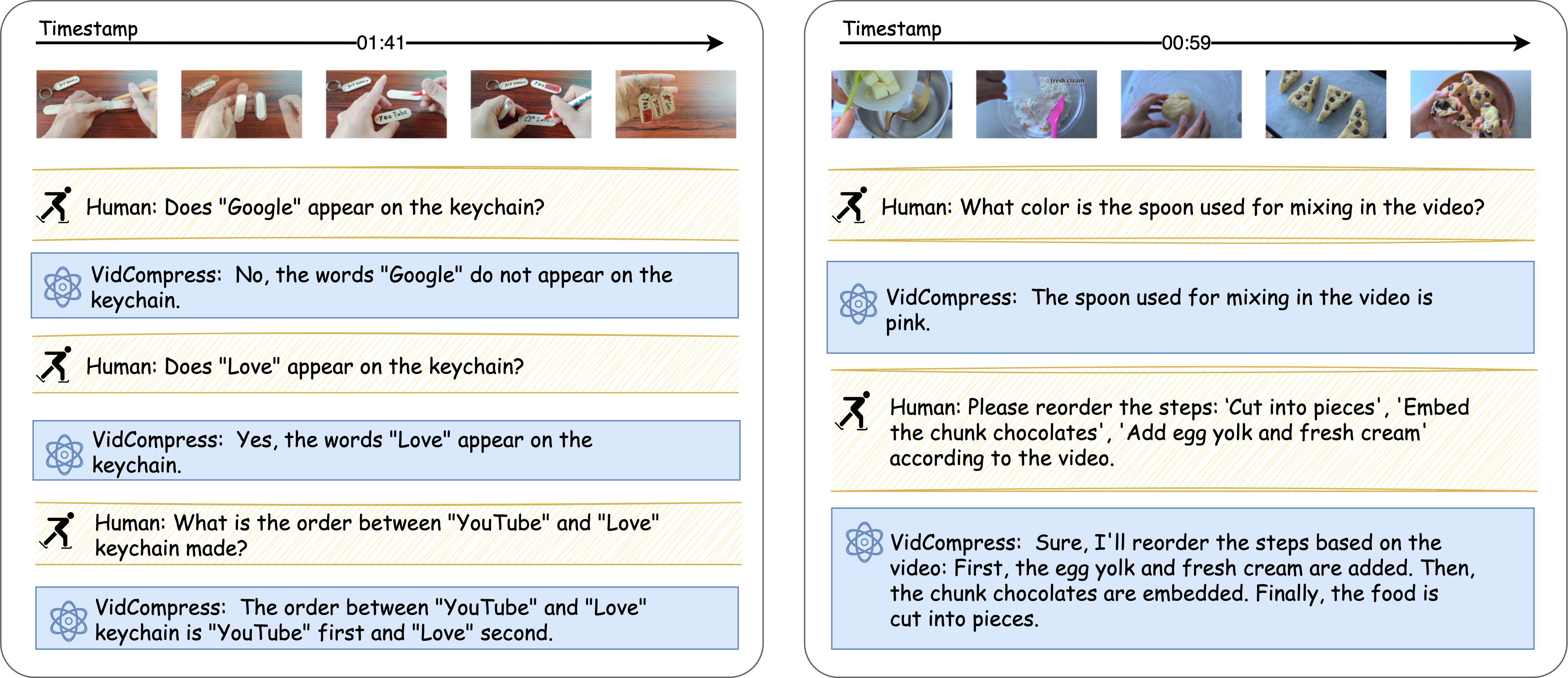}
	\caption{Chat examples of our VidCompress, with a DIY keychain video and a cooking video.
	}
	\label{fig:qua}
\end{figure*}
In this section, we conduct ablation studies by evaluating the effectiveness of the two different branches in VidCompress, with the results shown in Table~\ref{tab:ablation}. The settings of the ablation models are as follows:

\begin{itemize}
    \item VidCompress$_{\rm{mem}}$: This model retains only the memory-enhanced compressor branch and feeds only the memory-compressed tokens as visual tokens into the LLM.
    \item VidCompress$_{\rm{txt}}$: This model retains only the text-perceived compressor branch, removes the cross-attention module, and directly feeds the 32 query tokens produced by the Q-Former into the LLM after average pooling.
    \item VidCompress$_{\rm{full}}$: Our full model, VidCompress.
\end{itemize}

VidCompress$_{\rm{mem}}$ achieves the lowest results among these three models. This outcome is expected because, unlike the Q-former in the text-perceived compressor, our memory-enhanced compressor is not pretrained and is merely trained from scratch. Consequently, it is more challenging for it to achieve better performance with a limited training dataset. VidCompress$_{\rm{txt}}$ is a more general pipeline similar to previous Video-LLMs with Q-Former as the vision-text adapter, which achieves good results.  However, the role of the memory single-branch in the overall model should not be underestimated. When we add the memory-enhanced compressor branch to VidCompress$_{\rm{txt}}$ and feed both memory tokens and perceived tokens into the LLM, our full model, VidCompress$_{\rm{full}}$ shows significant improvement on MVBench and noticeable improvement on long video results in Video-MME. This indicates that the introduction of long-term memory information enhances the performance of Video-LLMs in analyzing video sequences.

\subsection{Ablation Studies on the Memory-Enhanced Compressor}

\begin{figure}[!b]
  \centering
  
  \subfloat[]{\includegraphics[width=0.9\linewidth]{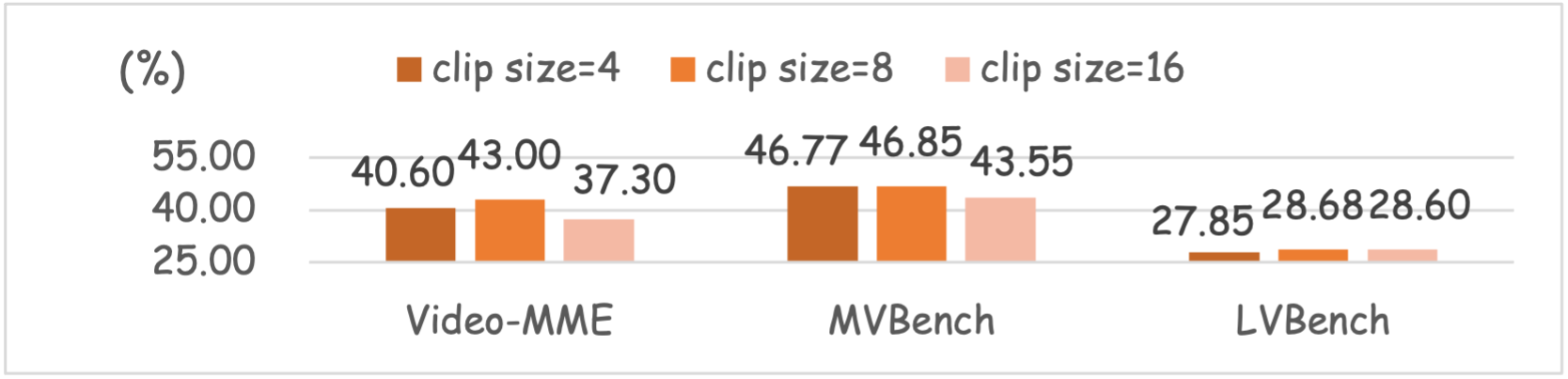}%
    \label{fig:ab_clip_size}} 
  
  \subfloat[]{\includegraphics[width=0.9\linewidth]{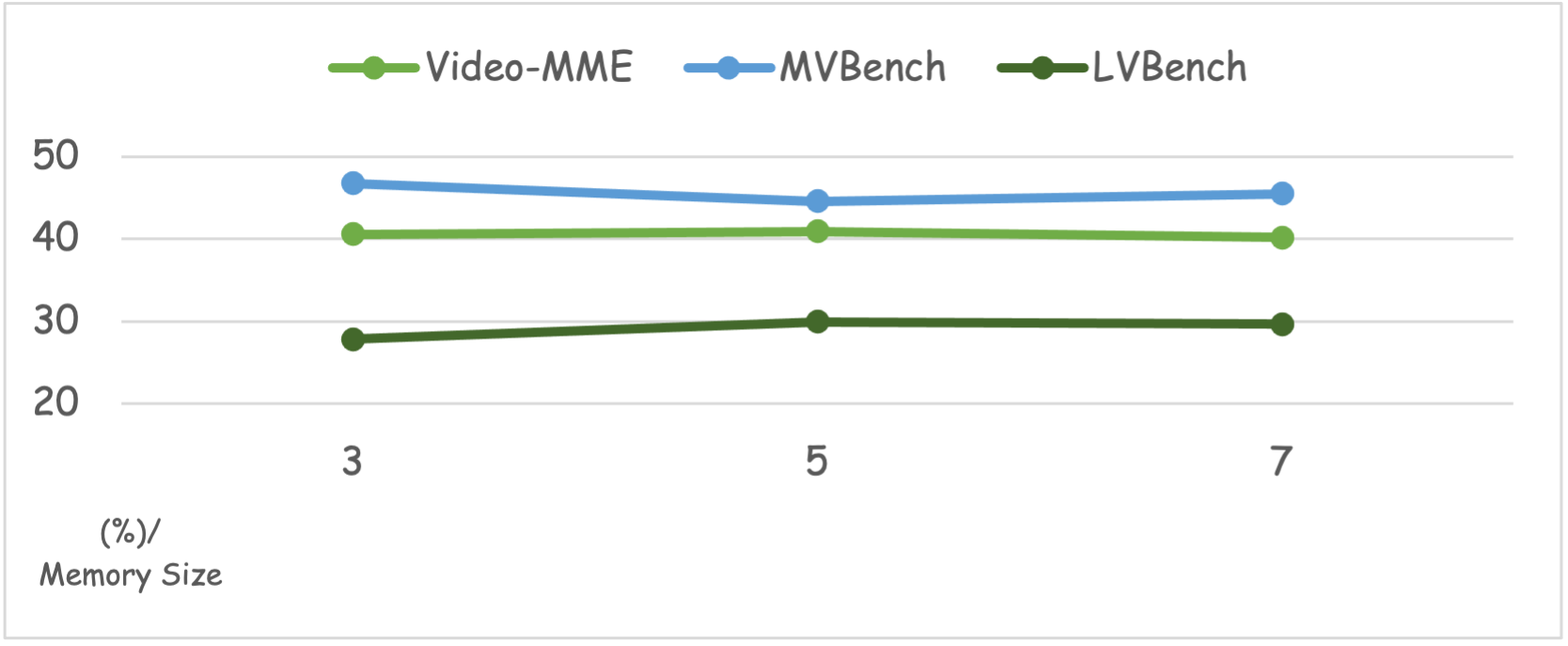}%
    \label{fig:ab_mem_size}}

  \caption{Ablation studies on (a) \textit{clip size} and (b) \textit{cached memory size}.}
  \label{fig:ab}
\end{figure}

The video clip size and cached memory size are two crucial factors that affect the memory-enhanced compressor, determining how far it can contact previous video sequences. Therefore, in this section, we conduct ablation studies on the video clip size and cached memory size to analyze their impacts on model performance (\cf, Figure~\ref{fig:ab}).

Figure~\ref{fig:ab_clip_size} presents the results of choosing different clip sizes (4, 8, and 16) on various video benchmarks, including Video-MME, MVBench, and LVBench. The results indicate a consistent trend regarding the impact of the clip size, \ie, across all three benchmarks, clip size 8 consistently delivers the best performance. While smaller clip size of 4 might lead to faster processing, the clip size of 8 significantly surpasses it on all three benchmarks. However, a larger clip size of 16 results in both performance drop on Video-MME and MVBench, surprisingly. Therefore, considering both efficiency and performance, clip size of 8 offers an optimal solution for VidCompress to capture both short-/long-term temporal relations for better video understanding.

We also investigate the impact of different cached memory sizes (3, 5, and 7) on the model performance, where the clip size is uniformly set to 4. As illustrated in Figure~\ref{fig:ab_mem_size}, the optimal size differs among those three benchmarks, \eg, as for MVBench the memory size should be 3 while the highest performance is observed with a memory size of 5 in terms of Video-MME. Also, the trend looks inconsistent across all three benchmarks. Given that increasing the memory cache size does not lead to significant improvements, it is most efficient to select the smallest memory cache size of 3 for the final solution, which needs less computation resource while maintaining comparable performance.

\subsection{Qualitative Results}
The qualitative results depicted in the Figure~\ref{fig:qua} showcase the capabilities of our proposed VidCompress model in understanding video content with fine-grained details and temporal relations. VidCompress accurately identifies and distinguishes textual information within videos, as demonstrated by its correct responses to questions regarding the presence of specific words (``Google'' and ``Love'') on a keychain. Additionally, the model effectively discerns visual details, such as identifying the color of a spoon used for mixing in a cooking video. Beyond these basic comprehension tasks, VidCompress also excels in understanding and reasoning about sequences of events, as evidenced by its ability to correctly reorder steps in a recipe according to the instructional video. These results highlight the model's strength in both fine-grained visual recognition and more complex temporal reasoning.


\section{Conclusion}
In this paper, we introduced VidCompress, a novel Video-LLM that addresses the challenges of temporal modeling in videos. VidCompress employs a dual-compressor architecture, combining a memory-enhanced compressor and a text-perceived compressor to generate two types of visual tokens. This design enhances both short-term and long-term temporal relationships and ensures fine-grained perception. Experiments on various video benchmarks and video question answering datasets demonstrate the superior performance of VidCompress, highlighting its ability to process and understand lengthy and complex video sequences. We believe VidCompress offers valuable insights for future research in enhancing temporal interactions in Video-LLMs.

\bibliography{reference}

\end{document}